\documentclass[conference]{IEEEtran}
\IEEEoverridecommandlockouts
\usepackage{cite}
\usepackage{amsmath,amssymb,amsfonts}
\usepackage{algorithmic}
\usepackage{graphicx}
\usepackage{textcomp}
\usepackage{xcolor}

\usepackage{graphicx}
\usepackage{xcolor,colortbl}
\usepackage[acronym]{glossaries}
\usepackage{amssymb} % For the checkmark
\usepackage{cleveref}
\usepackage{subcaption}
\usepackage{mwe}

\usepackage{tikz}
\usepackage{circledsteps}

\usepackage{booktabs}
\usepackage{multirow}
\usepackage{makecell}
\usepackage{caption}

\def\BibTeX{{\rm B\kern-.05em{\sc i\kern-.025em b}\kern-.08em
    T\kern-.1667em\lower.7ex\hbox{E}\kern-.125emX}}
\begin{document}

\title{Demystifying Drug Repurposing Domain Comprehension with Knowledge Graph Embedding
}

\author{
\IEEEauthorblockN{
Edoardo Ramalli\IEEEauthorrefmark{1},
Alberto Parravicini\IEEEauthorrefmark{1}, 
Guido W. Di Donato\IEEEauthorrefmark{1}, 
Mirko Salaris\IEEEauthorrefmark{1}\\
Céline Hudelot\IEEEauthorrefmark{2},
Marco D. Santambrogio\IEEEauthorrefmark{1}
}
\IEEEauthorblockA{\IEEEauthorrefmark{1}{Politecnico di Milano, DEIB, Milan, Italy},
\IEEEauthorrefmark{2}{Université Paris-Saclay CentraleSupeléc, MICS Lab, Gif-sur-Yvette, France}
}
\IEEEauthorrefmark{1}{\{edoardo.ramalli, alberto.parravicini, guidowalter.didonato, mirko.salaris, marco.santambrogio\}}@polimi.it\\
\IEEEauthorrefmark{2}{celine.hudelot@centralesupelec.fr}
}

\maketitle

\newacronym{kg}{KG}{Knowledge Graph}
\newacronym{kge}{KGE}{Knowledge Graph Embedding}
\newacronym{ner}{NER}{Named Entity Recognition}
\newacronym{hp}{HP}{Hyper Parameter}
\newacronym{dr}{DR}{Drug Repurposing}
\newacronym{ml}{ML}{Machine Learning}
\newacronym{gml}{GML}{Graph Machine Learning}
\newacronym{snap}{SNAP}{Stanford Network Analysis Platform}
\newacronym{hits}{H@N}{Hits@N}
\newacronym{biokg}{BioKG}{\texttt{ogbl-biokg}}
\newacronym{lp}{LP}{Link Prediction}

%%%%%%%%%%%%%%%%%%%%%%%%%%%%%%%%%%%%%%%%%%%%%%%%%%%%%%%%%%%%%%%%%%%%
\begin{abstract}
Drug repurposing is more relevant than ever due to drug development's rising costs and the need to respond to emerging diseases quickly. Knowledge graph embedding enables drug repurposing using heterogeneous data sources combined with state-of-the-art machine learning models to predict new drug-disease links in the knowledge graph. As in many machine learning applications, significant work is still required to understand the predictive models' behavior. We propose a structured methodology to understand better machine learning models' results for drug repurposing, suggesting key elements of the knowledge graph to improve predictions while saving computational resources. We reduce the training set of 11.05\% and the embedding space by 31.87\%, with only a 2\% accuracy reduction, and increase accuracy by 60\% on the open ogbl-biokg graph adding only 1.53\% new triples.
%Drug repurposing is more relevant than ever due to drug development's rising costs and the need to respond to emerging diseases quickly. Knowledge graph embedding enables drug repurposing using heterogeneous data sources combined with state-of-art machine learning models to predict new drug-disease links in the knowledge graph. As in many machine learning applications, significant work is still required to understand the predictive models' behavior. We propose a structured methodology to better understand the results of machine learning models for drug repurposing, suggesting key elements of the knowledge graph to improve predictions while saving computational resources. We reduce the training set of 11.05\% and the embedding space by 31.87\% with only a 2\% reduction in accuracy, and increases the accuracy by 60\% on the open ogbl-biokg graph by adding only 1.53\% new triples.
\end{abstract}

\begin{IEEEkeywords}
Drug Repurposing, Biomedical Knowledge Graph, Knowledge Graph Embedding, Link Prediction, Machine Learning
\end{IEEEkeywords}

\section{INTRODUCTION}\label{sec:intro}
\begin{figure*}[t]
      \centering
      \includegraphics[width=\textwidth]{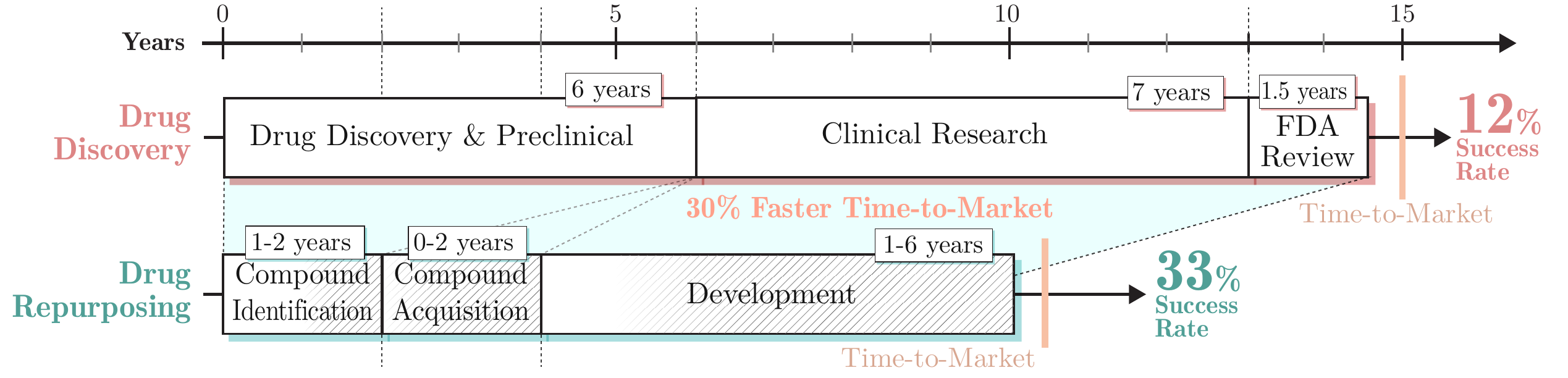}
      \caption{Drug Discovery and Drug Repurposing timeline.}
      \label{fig:timeline}
      \vspace{-10pt}
\end{figure*}

Discovering new drugs is a tricky, expensive, and slow mission. It involves different stages that often require clinical trials to move forward. \gls{dr} discovers new therapeutic uses for existing drugs, reducing time-to-market by 30\% to 80\% (\Cref{fig:timeline}) and cost ($\sim$80\%) with a lower failure risk in the trails compared to a new chemical entity \cite{pushpakom2019drug}. In the last years, the number of available biomedical information increased to the point that it is arguably impossible to manage it manually \cite{BigData}. Fortunately, automatic procedures significantly benefit from integrating heterogeneous information from different sources of data. This challenge has spurred significant research in Biomedical Informatics, intersecting disciplines such as data integration and representation learning. 
While results are promising, formal methodologies taking into account biomedical knowledge are essential to better understand these outcomes \cite{BiomedicalInformatics,795}. 
% We applied \gls{ml} to \gls{dr}, and investigate the meaning behind the prediction scores for different analyses, ending up proposing a methodological approach to understand the influence of the network structure, used to represent a biomedical domain, on the prediction accuracy for graph representation learning applications.
We applied \gls{ml} to \gls{dr}, and, to achieve higher interpretability, we investigated the meaning behind the prediction scores with different analyses.
We propose a methodology to understand how the network structure representing a biomedical domain influences the prediction accuracy in graph representation learning applications. As a result, we can leverage this information to improve the quality of the network and subsequently improve the predictions and reduce the computational resources needed to learn the representation of the biomedical domain.

A \gls{kg} is a network of heterogeneous entities connected by specific relationships capable of representing a complex domain semantic \cite{795}. Encoding the biomedical domain in a \gls{kg} translates the task of \gls{dr} into finding possible new connections between a drug and a disease \cite{gaudelet2021utilising}. 
Reusing a pre-existent biomedical \gls{kg} is difficult because databases often restrict data redistribution without a commercial license. We represent a biomedical domain that is beneficial for \gls{dr}, combining different available sources of information. We leverage different representation learning techniques to extract knowledge from structured and unstructured data from curated databases such as Uniprot \cite{uniprot}, CTD \cite{ctd}, DrugBank \cite{drugbank}, and OMIM \cite{omim}, pursuing the importance of data sharing in the biomedical field \cite{reuse}.

\Glspl{kge} are representations obtained through \gls{ml} techniques that project the \gls{kg} to a lower-dimensional space that preserves the graph structure \cite{Nickel_2016}.
\Glspl{kge} can predict new relationships between entities in the graph: in the context of \gls{dr}, we can leverage embeddings to discover new links between drugs and diseases.

Transparency, interpretability, and explainability are long-standing issues in the application of \gls{ml} to natural sciences \cite{ExplainableML}.
For this reason, we studied the quality of \gls{dr} predictions obtained from \glspl{kge} with different types of analysis, combining domain knowledge with the \gls{ml} outcomes. 
Our methodology also hints at strategies to reduce the computational cost of \glspl{kge}, with insignificant accuracy detriment.

% The interpretation of the results in the biomedical domain knowledge underlines the importance of graph structure in \glspl{kge} to enhance the embedding quality. 

%Vogliamo dire due cose:
%- La struttura del grafo è importante per KGE
%- In un dataset biomedico puoi interpretare meglio i %risultati

%La struttura del grafo è importante perchè in un dominio biomedico le entità sono 'connesse' e di conseguenza i risultati hanno una maggiore interpretability.

This work can be valuable to the pharmaceutical industry, having the potential to speed up the \textit{compound identification} stage of the \gls{dr} approach, which can require up to 2 years of work (\Cref{fig:timeline}). Moreover, our methodology reduces the \gls{kge} representation size on a novel \gls{kg} by 31.87\% (with a comparable reduction in training time), with only a 2\% accuracy reduction, and increases the accuracy by 60\% on the open \gls{biokg} graph \cite{ogb} with the addition of only 1.53\% new triples.

%In summary, we present the following contributions:
Our main contributions are:
\begin{itemize}
    \item A new biomedical \gls{kg} built from free databases specifically to address \gls{dr} (\Cref{sec:ourkg}).
    \item A methodology to analyze the efficacy of embedding models for \gls{dr} while saving resources (\Cref{sec:methodology}).
    \item An analysis of the relationship between the prediction quality and data structure, discovering which links and entities are more incisive in \gls{dr} (\Cref{sec:evaluation}).
\end{itemize}

\section{DATA \& METHODS}
The literature offers several embedding models for \glspl{kg} \cite{rossi2020knowledge} and describes various biomedical databases used for different applications, in particular to apply \gls{gml} techniques to \gls{dr} \cite{gaudelet2021utilising, NICHOLSON20201414}. 

This section introduces the general idea of the embedding model used to support \gls{dr} and presents the used datasets.

\subsection{Knowledge Graph and Knowledge Graph Embedding}
\begin{figure}[t]
      \centering
      \includegraphics[width=\columnwidth]{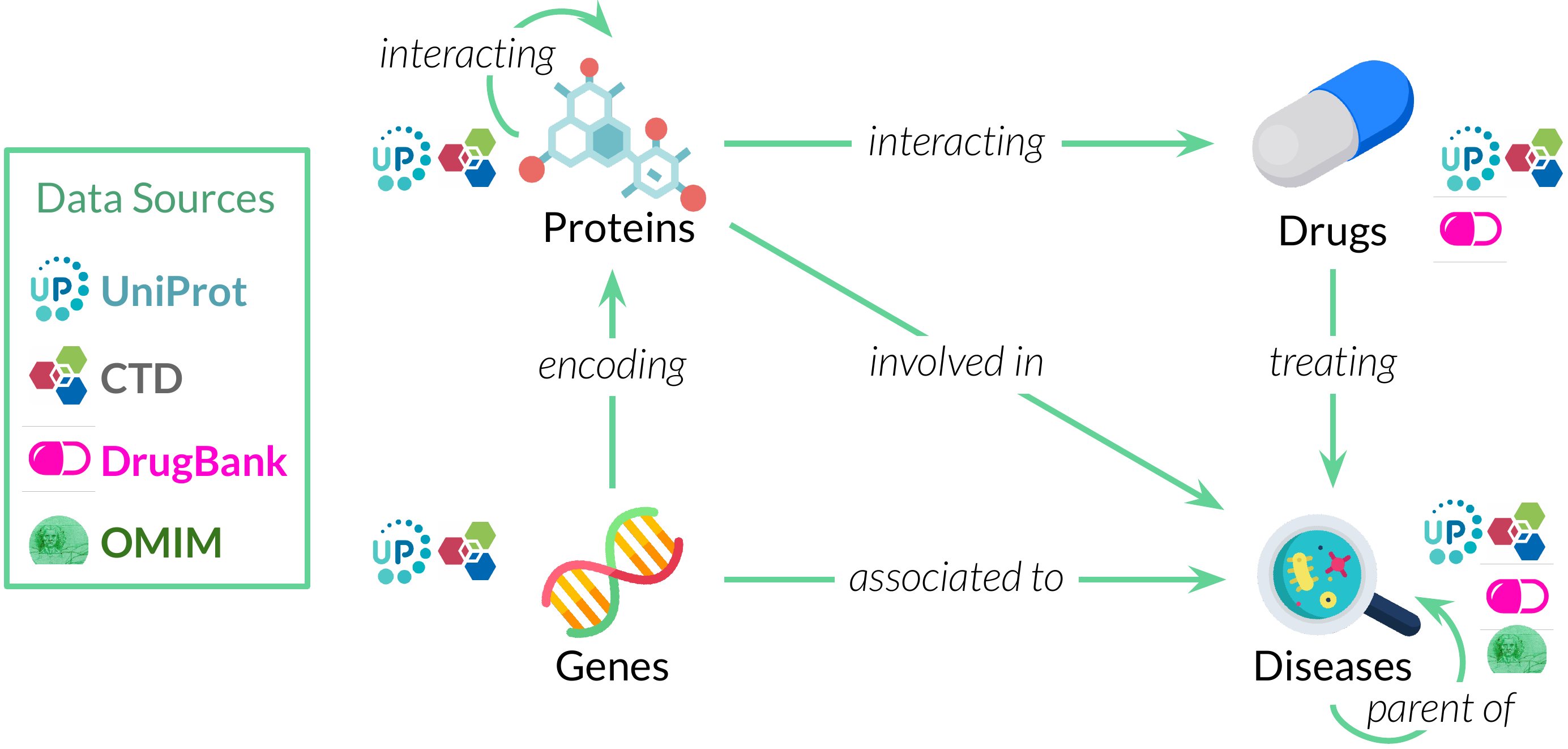}
      \caption{The schema of our biomedical \gls{kg} with four different types of entities and specific relationships connecting them.}
      \label{fig:ourkg}
    %   \vspace{-10pt}
\end{figure}

A \gls{kg} is a network that specifies the type of connection between two entities \cite{definition}: \Cref{fig:ourkg} represents the schema of the \gls{kg} built for this work. \glspl{kg} are commonly represented as lists of triples. A triple is composed of three elements: two entities, called head $h$ and tail $t$, and a relationship $r$ that connects them. Due to the \gls{kg} structure, it is easy to integrate the graph with heterogeneous information sources.

\gls{kge} models represent the \gls{kg} in a lower-dimensional space that condenses and preserves the original information but also allows extracting hidden information. A \gls{kge} differs from another by the representation space, the scoring function, and the encoding models' additional features \cite{rossi2020knowledge}.
The embedding model uses a representation space with a mathematical structure to encode peculiar relation properties of the \gls{kg} into a low-dimension representation vector. The model score function measures the embedded triple's plausibility together with additional information from the graph.

\gls{lp} is the task of predicting facts in a \gls{kg} to forecast the existence of a missing triple, leveraging the learned embedded representation \cite{rossi2020knowledge}. As such, \gls{dr} can be seen as \gls{lp} between a drug and a disease.

\subsection{Selecting an Embedding Model}
Multiple state-of-the-art \gls{kge} techniques exist, without a single one outperforming the others in similar conditions \cite{rossi2020knowledge, ampligraph}. For our application, we found that TransE is the best compromise between computational complexity and prediction accuracy. 
TransE requires fewer data and parameters than competing embedding models to provide excellent accuracy \cite{rossi2020knowledge}; moreover, it can also scale to more extensive databases, making it suitable for easy prototyping.
TransE models relationships by interpreting them as translations operating on the entities' low-dimensional embeddings \cite{TransE}.
Given a training set $\mathbf{S}$ of triples $\{(h, r, t) \in \mathbf{S}\}$, TransE learns for each entity $h$ and $t$, and for each relation $r$, a vector representation of chosen size $K$.
The primary idea behind TransE is that the functional relation induced by the $r$-labeled edges corresponds to a translation of the embeddings. The vector sum of $h + r = t$ when $(h, r, t)$ is a true triple ($t$ should be the nearest neighbor of $h + r$), while $h + r$ should be far away from $t$ if the triple is false, i.e. a negative triple, denoted as  $(h', r, t') \in \mathbf{S'}$, with $S' \cap S = \emptyset$.
TransE, to learn such embedding, minimizes a loss function $\mathcal{L}$ (Eq. \ref{eq:loss_transe}), computed as sum of dissimilarity measure $d$ (L1 or L2 norm) over the training set \cite{TransE}.
\begin{equation} \label{eq:loss_transe}
   \mathcal{L} = \sum_{(h, r, t) \in \mathbf{S} }{\sum_{(h', r, t') \in \mathbf{S'}}{[d(h+r, t) - d(h' + r, t')]_{+}}}
\end{equation}

To predict a new link in the graph, the model replaces a triple's element with a specific subset of entities, and it ranks each new triple using the cost function against the learned embedding. The top-ranking triples are plausible triples, and for this reason, possible new connections in the \gls{kg} \cite{TransE}.

\subsection{Integrating Heterogeneous Sources of Data}
\label{sec:ourkg}

To represent the heterogeneous information of a biomedical domain as a \gls{kg}, it is necessary to start from curated databases \cite{NICHOLSON20201414}.
To address \gls{dr}, we propose a \gls{kg} which is the result of the combination of heterogeneous sources of information. We combine free biomedical curated databases composed of unstructured information as in DrugBank \cite{drugbank} and UniProt \cite{uniprot}, and structured information as in CTD \cite{ctd}, and OMIM \cite{omim}.
In the case of unstructured information fields, we use the biomedical \texttt{en\_ner\_bc5cdr\_md} \gls{ner} system \cite{neumann-etal-2019-scispacy} to extract meaningful connections between entities from the textual content.
Although the \gls{kg} is easily extensible, the integration process is not immediate due to the naming complexity of the different entities. The same entity can have more commercial names, or a database can use a different naming system that made it necessary to use dictionary databases, like OMIM, to homogenize diverse names. The result is a compact, effective \gls{kg} built to address the problem of \gls{dr}.
We use a second biomedical \gls{kg}, \gls{biokg} \cite{ogb}. This graph contains more entities and relations than ours, but it is not built to specifically target \gls{dr}.
In \Cref{tab:kg_numbers}, there are the characteristics of the two biomedical \gls{kg}. In particular, we observe that our \gls{kg} has fewer triples overall, but it has more drug-disease relationships that are useful for \gls{dr}. This consideration holds true even if \gls{biokg} has other drugs-related information such as side effect entities.

\newcommand*\rot{\rotatebox{0}}
\renewcommand\theadalign{tc}
\renewcommand\theadfont{\bfseries}
\setlength\tabcolsep{5pt}
% \renewcommand\theadgape{\Gape[10pt]}
% \renewcommand\cellgape{\Gape[10pt]}

% \begin{table}
% \centering
% \renewcommand{\arraystretch}{1.1} 
%     \caption{Number of entities/relations (and percentage of total) in each \glspl{kg} in our evaluation. Highest values in bold.}
%     \resizebox{\columnwidth}{!}{
% 	\begin{tabular}{@{}llcccc@{}}
% 		\toprule
% 	    & \phantom{} & \thead{Diseases} & \thead{Drugs}  & \thead{Genes}  & \thead{Proteins} \\
% 		\cmidrule{3-6}
% 		\textbf{Our \gls{kg}} && 4596 (7.32\%)  &\textbf{13945 (22.21\%)} & \textbf{20017 (31.87\%)} &\textbf{24242 (38.6\%)} \\ 
%         \textbf{ogb-biokg} &&\textbf{10687 (11.20\%)} & 10533 (11.23\%) & 0 & 17499 (18.66\%)\\
%     	\addlinespace
%     	& \phantom{} & \thead{Side-Effects} & \thead{Functions}  & \thead{Relations}  & \thead{Triples} \\
%     	\cmidrule{3-6}
%     	\textbf{Our \gls{kg}} && 0 & 0 & \textbf{69} & 183000 \\ 
%         \textbf{ogb-biokg} && \textbf{9969 (10.63\%)} & \textbf{45085 (48.08\%)} & 51 & \textbf{4760000} \\
% 		\bottomrule
% 	\end{tabular}
%     }
%     \label{tab:kg_numbers}
% \end{table}

\begin{table*}
\centering
\renewcommand{\arraystretch}{1.} 
    \caption{Number of entities/relations (and percentage of total) in each \glspl{kg} in our evaluation. Highest values in bold.}
    \resizebox{\textwidth}{!}{
	\begin{tabular}{@{}lcccccccccccc@{}}
		\toprule
	    \phantom{} &\multirow{2}{*}{\rot{\thead{Diseases}}} & \multirow{2}{*}{\rot{\thead{Drugs}}}  & 
	    \multirow{2}{*}{\rot{\thead{Genes}}}  & 
	    \multirow{2}{*}{\rot{\thead{Proteins}}} & 
	    \rot{\thead{\makecell{Side\\Effects}}} & 
	    \multirow{2}{*}{\rot{\thead{Functions}}}  & 
	    \rot{\thead{\makecell{Total\\Entities}}} & 
	    \phantom{}&
	    \rot{\thead{Drug-Disease\\Triples}} & 
	    \rot{\thead{Drug-Drug\\Triples}}  &  
	    \rot{\thead{Disease-Disease\\Triples}}  & 
	    \rot{\thead{\makecell{Total\\Triples}}}\\
		\cmidrule{2-8} \cmidrule{10-13}
    		
    		\makecell[l]{\textbf{Our \gls{kg}}} 
    		&
    		\makecell{4596\\ (7.32\%)}
    		&
    		\makecell{\textbf{13945}\\ \textbf{(22.21\%)}}
    		&
    		\makecell{\textbf{20017}\\ \textbf{(31.87\%)}}
    		&
    		\makecell{\textbf{24242}\\ \textbf{(38.6\%)}}
		    &
        	0
        	&
        	0
        	&
        	62800&
        	&
        	\makecell{\textbf{72976}\\\textbf{(39.9\%)}}
        	&
        	0
        	&
        	\makecell{\textbf{21913}\\\textbf{(11.97\%)}}
        	&
        	183000
		\\ 
		\addlinespace
            \makecell[l]{\textbf{ogb-biokg}}
            &
            \makecell{\textbf{10687}\\\textbf{(11.20\%)}}
            &
            \makecell{10533\\ (11.23\%)}
            &
            0
            &
            \makecell{17499\\ (18.66\%)}
             &
            \makecell{\textbf{9969}\\ \textbf{(10.63\%)}}
            &
            \makecell{\textbf{45085}\\ \textbf{(48.08\%)}}
            &
            \textbf{93773}&
            &
            \makecell{5147\\($<1\%$)}
            &
            \makecell{\textbf{1133686}\\ \textbf{(23.82\%)}}
            &
            0
            & \textbf{4760000}
        \\
		\bottomrule
	\end{tabular}
    }
    \label{tab:kg_numbers}
    \vspace{-5pt}
\end{table*}

% \begin{table}[]
% \begin{tabular}{|l|c|c|c|c|}
% \hline
% \cellcolor[HTML]{C0C0C0} & \multicolumn{1}{l|}{\textbf{Diseases}}     & \multicolumn{1}{l|}{\textbf{Drugs}}     & \multicolumn{1}{l|}{\textbf{Genes}}     & \multicolumn{1}{l|}{\textbf{Proteins}} \\ \hline
% \textbf{Our KG}          & 5k                                         & 13k                                     & 21k                                     & 23k                                    \\ \hline
% \textbf{ogb-biokg}       & 11k                                        & 11k                                     & 0                                       & 18k                                    \\ \hline
% \cellcolor[HTML]{C0C0C0} & \multicolumn{1}{l|}{\textbf{Side-Effects}} & \multicolumn{1}{l|}{\textbf{Functions}} & \multicolumn{1}{l|}{\textbf{Relations}} & \multicolumn{1}{l|}{\textbf{Triples}}  \\ \hline
% \textbf{Our KG}          & 0                                         & 0                                       & 69                                      & 183k                                   \\ \hline
% \textbf{ogb-biokg}             & 10k                                        & 45k                                     & 51                                      & ?                                      \\ \hline
% \end{tabular}
% \label{tab:kginfo}
% \caption{Dettagli}
% \end{table}

\subsection{Proposed analysis methodology}
\label{sec:methodology}
We propose the following procedure to analyze the embedding applied on \gls{kg} as a result of the experimental results in \Cref{sec:evaluation}: \Circled{1} Set the embedding model with \glspl{hp} coming from a \gls{kg} with similar size and domain. This step helps to reduce the \glspl{hp} research space.  \Circled{2} Optimization of \glspl{hp}, in such a way that the fine-tuning of the embedding model produces the best prediction performances. \Circled{3} Carry out feature ablation or extension analyses that determine if the model is learning and not memorizing. This analysis aims to highlight the \gls{kg} semantic and structural strengths and criticalities. Leveraging this information, we can understand if there are not such helpful parts of the \gls{kg} but that have a non-negligible impact on the resources used for the training procedure.

\section{EXPERIMENTAL EVALUATION}
\label{sec:evaluation}

\begin{figure}
    \begin{subfigure}[t]{0.75\columnwidth}
        \centering
        \includegraphics[width=\columnwidth]{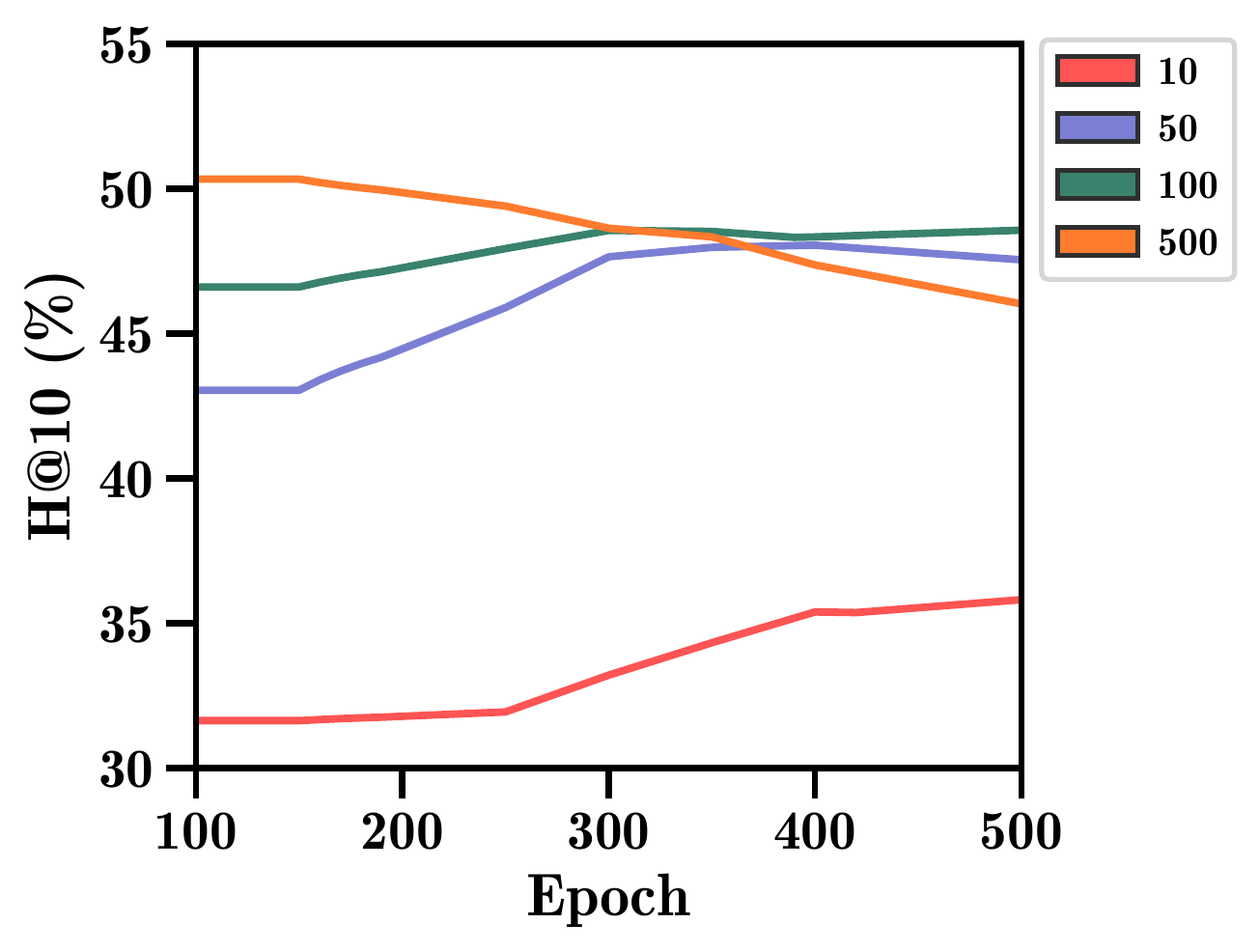}
        \caption[]%
        {{\small Embedding Dimension}}    
        \label{fig:embDim}
    \end{subfigure}
    \begin{subfigure}[t]{0.75\columnwidth}  
        \centering 
        \includegraphics[width=\columnwidth]{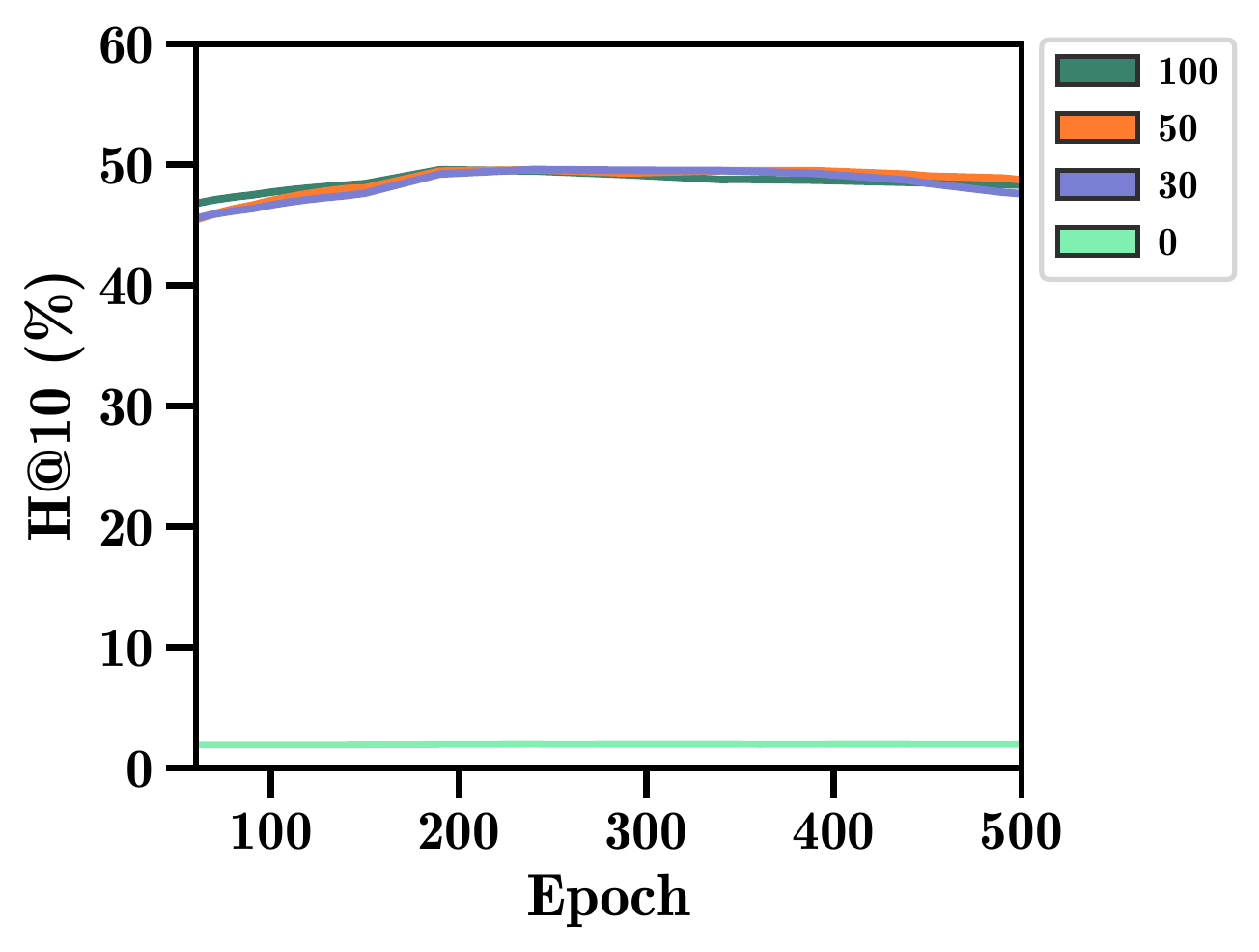}
        \caption[]%
        {{\small Negative Sampling}}    
        \label{fig:negative}
    \end{subfigure}
    \centering
    \caption{Importance of the embedding dimension and negative sampling for prediction score and computational resources.}
    \label{fig:HP2}
\end{figure}

\begin{figure}
    \begin{subfigure}[t]{0.75\columnwidth}   
        \centering 
        \includegraphics[width=\columnwidth]{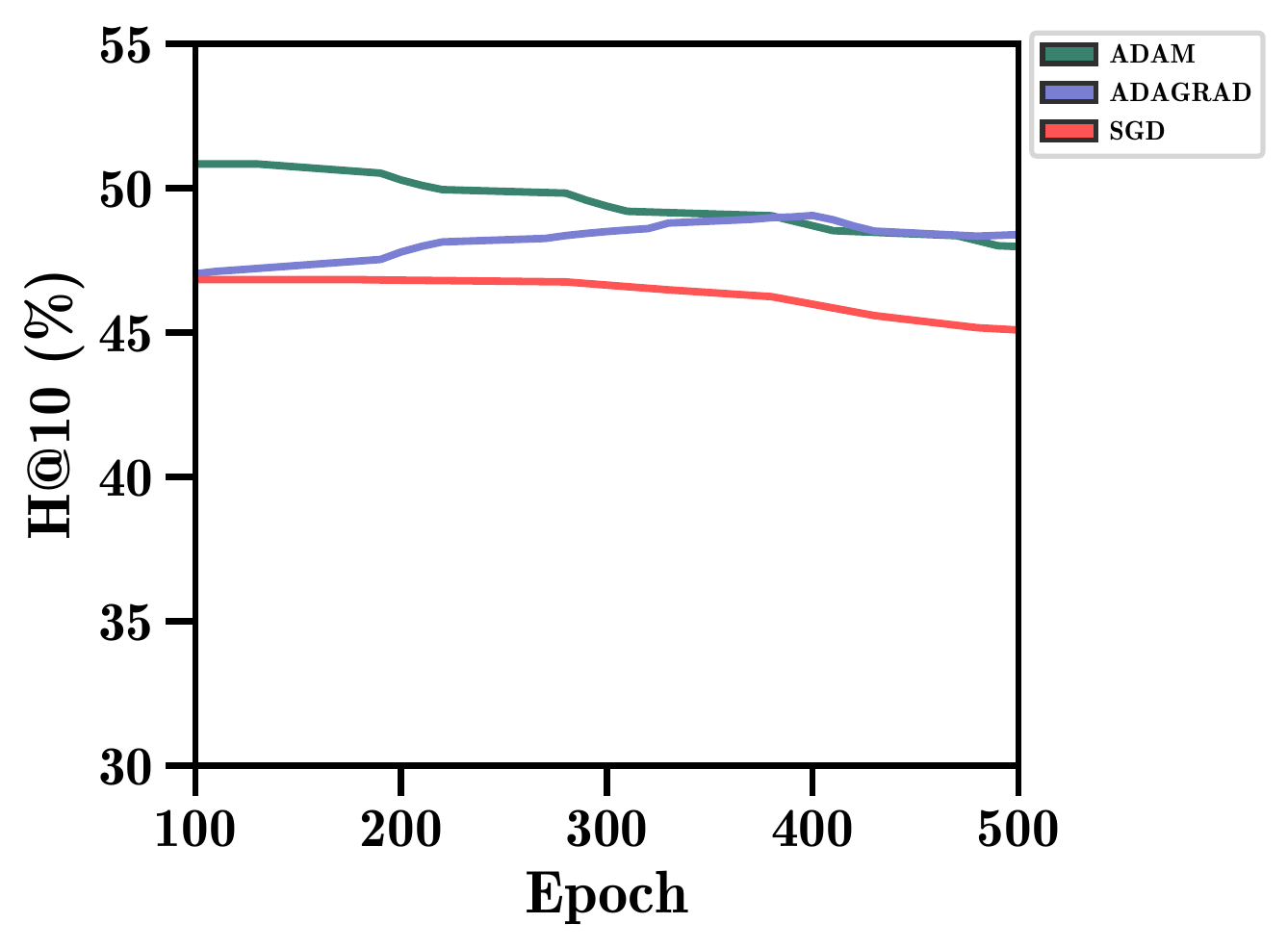}
        \caption[]%
        {{\small Optimizer}}    
        \label{fig:optimizer}
    \end{subfigure}
    \begin{subfigure}[t]{0.75\columnwidth}   
        \centering 
        \includegraphics[width=\columnwidth]{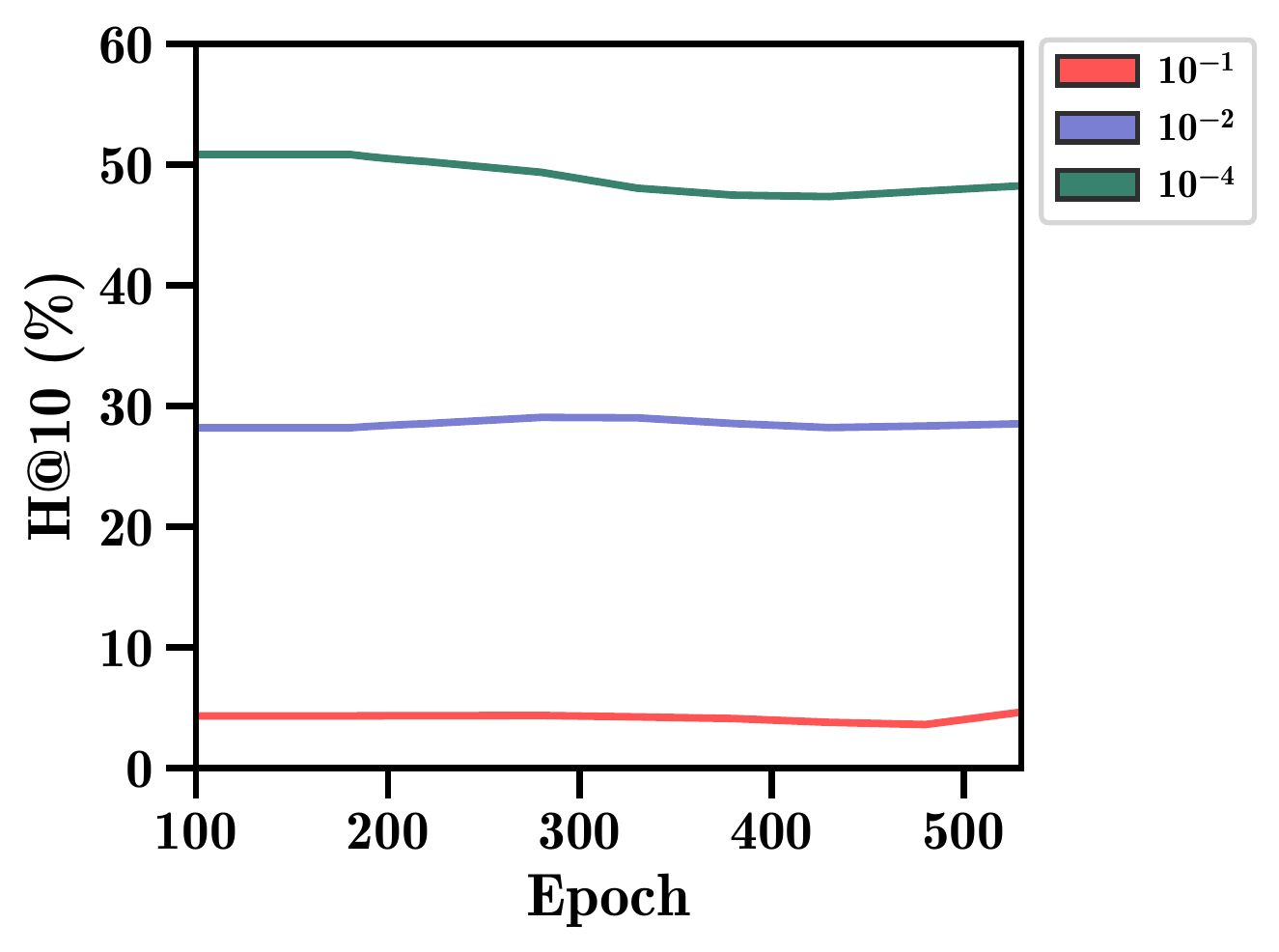}
        \caption[]%
        {{\small Learning Rate}}    
        \label{fig:lr}
    \end{subfigure}
    \centering
    \caption{Importance of the optimizer and learning rate for prediction score and computational resources.}
    \label{fig:HP2}
    % \vspace{-10pt}
\end{figure}

This section presents the study results on the model hyperparameters, then shows the prediction accuracy for \gls{dr} for both datasets, and finally investigates these results with feature ablation and graph extension techniques.
We randomly split drug-disease triples into training, validation, and test set with a probability of 60\%, 20\%, 20\%, respectively. Non-drug-disease triples are also added to the training set.
Focusing on the \gls{dr} task, we measure the embedding model's prediction accuracy using \gls{hits} only against the drug-disease triples present in the validation set.
\gls{hits} is the proportion of the original correct triples to the top $N$ predictions of the model (Eq. \ref{eq:hits@n}).
Other public results instead provide general results on all possible link predictions \cite{ogb}.
To compute this metric, we follow the procedure described in \cite{TransE,ogb}.
The process consists of corrupting the validation set's triples $Q$, by replacing one entity of the triple with another entity from a random subset of the same type.
The embedding model ranks all the corrupted triples and the valid triples against each other with \gls{hits}.
\begin{equation} \label{eq:hits@n}
    Hits@N =\frac{1}{|Q|} \sum_{i = 1}^{|Q|}  \begin{cases}
           1 & \text{if}\ rank_{(h,r,t)_i} \leq N\\
           0 & \text{Otherwise}
   \end{cases}
\end{equation}

We average \gls{hits} scores over 10 independent training/validation cycles, with negligible variance.

\subsection{Hyper Parameters}
The embedding model requires \glspl{hp} tuning to be trained effectively on a specific dataset.
The two most important \glspl{hp} in a \gls{kge}, according to \cite{rossi2020knowledge}, are the embedding dimension and the optimizer (with its learning rate).
Other parameters, like the negative sampling, can affect the time to train a model, but they are less significant for the final accuracy.
In particular, in \Cref{fig:negative}, negative sampling is difficult to manage since it is hard to have a negative set (a set of false triples) available. Perturbing the triples randomly is challenging as there is no certainty that this is not a possible repurposed drug, and inserting it in the negative set would indicate to the model to penalize an actually correct representation of the triple. For this reason, we choose a low value for the negative sampling that reduces the probability of this event and saves computational resources.
The result shows that a low embedding dimension yields the worst accuracy. Instead, an exaggerated embedding dimension does not bring benefits but only faster overfitting and a higher computational cost. For the \gls{kg} proposed in this work, the best embedding size is 128 (\Cref{fig:embDim}) since it is the best compromise between accuracy and model complexity. The best optimizer proved to be ADAM with learning rate $\lambda = 10^{-4}$, coherently to \cite{ampligraph} (\Cref{fig:optimizer,fig:lr}).

\subsection{Drug Repurposing Prediction Score}
\label{sec:generic}
TransE applied to our biomedical \gls{kg} achieves a \gls{hits} score slightly above $52\%$.
In other words, the model proposes a correct repurposing in the first ten predictions $52\%$ of the times. 
This result shows that our model has significant learning capabilities: a random baseline (10 random drugs chosen as repurposing candidates) always has \gls{hits} close to $0\%$ due to the enormous number of possible combinations.

\begin{figure}[t]
      \centering
      \includegraphics[width=0.85\columnwidth]{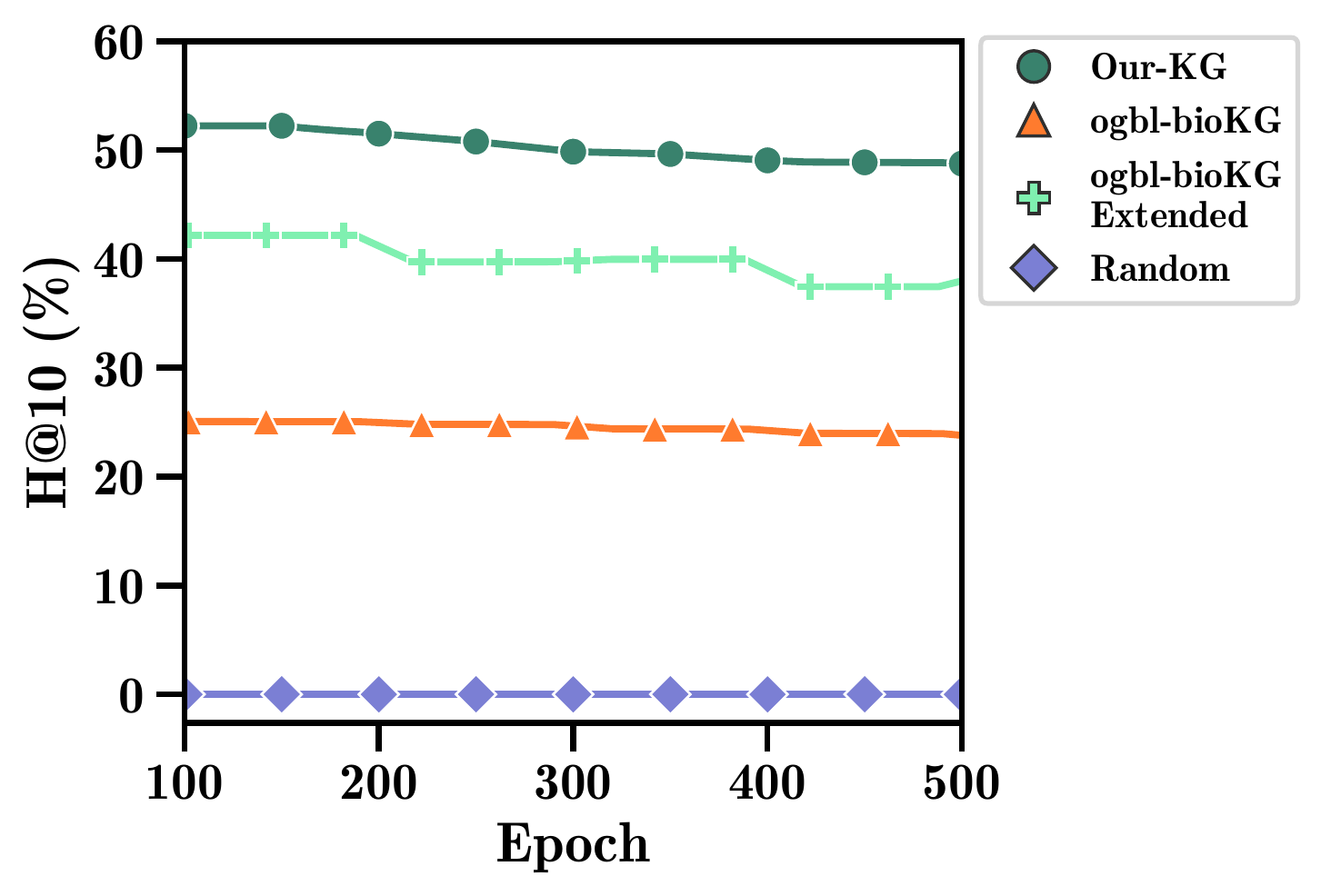}
      \caption{Prediction \gls{hits} score for TransE applied to our KG and to \gls{biokg}, compared to a random predictor.}
      \label{fig:generic}
    %   \vspace{-3pt}
\end{figure}

The same embedding algorithm applied to \gls{biokg} gives \gls{hits} accuracy below $30\%$, when predicting the same drug-disease relations. \gls{biokg} contains more triples than our \gls{kg} but less useful information for \gls{dr}.
From these promising results (\Cref{fig:generic}), we investigate how the \gls{kge} structure relates to such different accuracies, in the next section.

\subsection{Feature Ablation}
\begin{figure}[t]
      \centering
      \includegraphics[width=0.85\columnwidth]{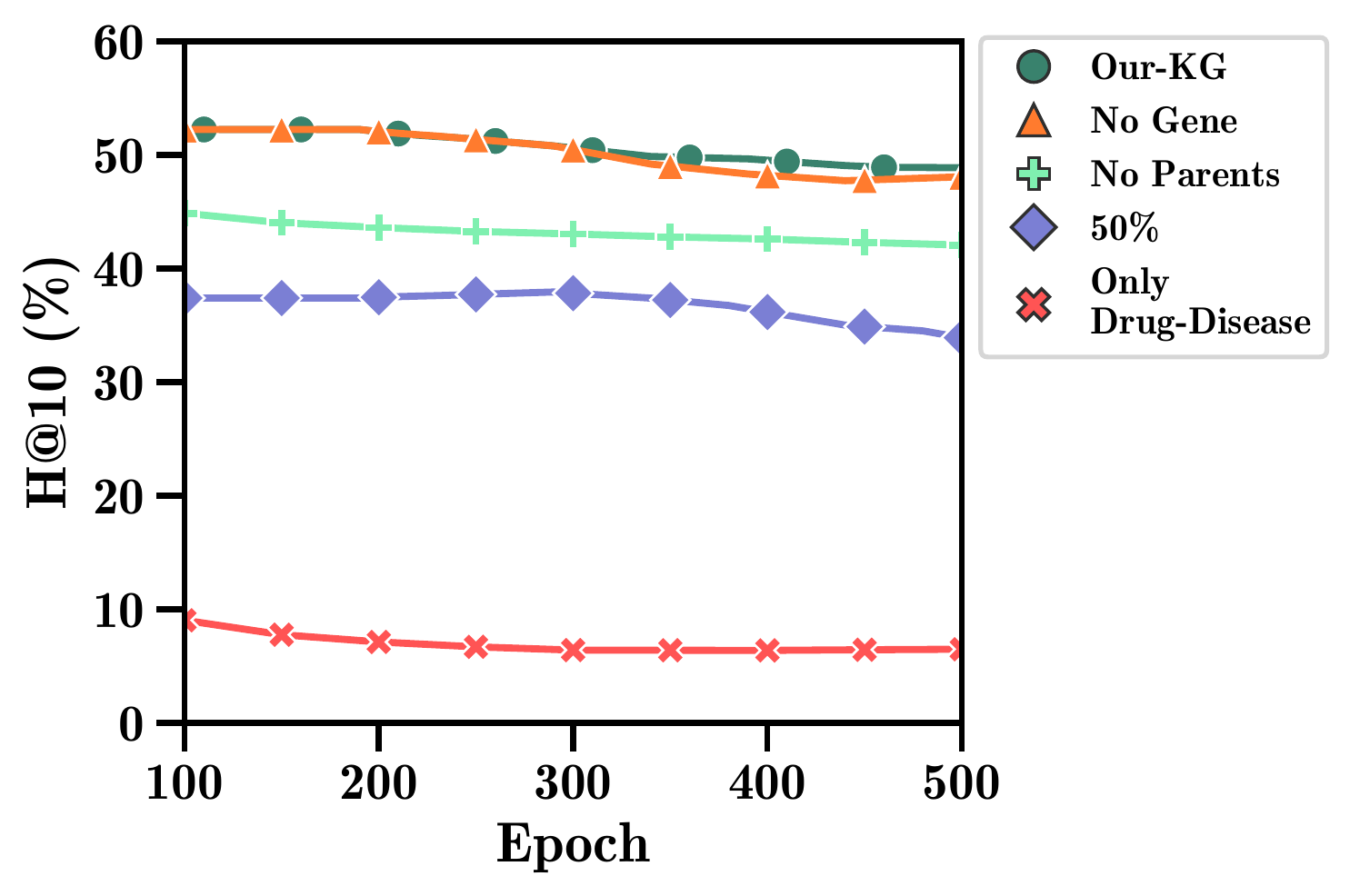}
      \caption{Comparing features ablation for H@10 accuracy.}
      \label{fig:ablation}
    %   \vspace{-8pt}
\end{figure}

To understand which parts of the input data are more critical in the training procedure, we systematically apply \textit{feature ablation} to our \gls{kg}, reducing its size and entity types. This methodology highlights which part of the \gls{kg} is critical to the \gls{dr} task and sheds a light on how the \gls{kge} model relates to the graph structure, the model outcomes, and the domain knowledge. A summary of the results is in \Cref{fig:ablation}.

\subsubsection{Only Drug-Disease}
If the dataset contains only drug-disease relationships, the accuracy score drops below $16\%$. This striking accuracy loss shows that the other removed triples are essential for good predictions.

\subsubsection{No gene}
Removing gene entities from the \gls{kg} results in a $\sim2\%$ \gls{hits} loss. Although gene entities represent more than $30\%$ of the entities in the \gls{kg}, they do not appear to be essential for \gls{dr}. This result suggests an important consequence: bigger \gls{kg}, with a higher computational cost for training and inferences, do not provide a significant improvement of \gls{dr} accuracy if they do not contain meaningful triples for the problem at hand.

\subsubsection{No Parents}
Another test is to remove from the dataset the relation that connects a disease to another one. This kind of relation expresses a hierarchy between diseases: a disease can be classified based on its specification, but it belongs to a more generic family group. This information can be helpful because it is very likely that a drug that treats a disease could benefit a similar drug that belongs to the same family.
This idea is confirmed by the results of the H@10 score on the validation set. If the disease-disease relations, representing 10\% of the dataset, are removed from the training set, accuracy decreases by 5-10\%.

\subsubsection{50\% Triples Removal}
Randomly removing 50\% of the triples has a clear impact on the accuracy result. The results show that increasing the number of information in the KG improves the model's accuracy significantly even if the score is not proportional to the graph's dimension.

\glsreset{biokg}

\subsection{Extension of \gls{biokg}}
Extending \gls{biokg} with other drug-disease relations used in our \gls{kg}, with the support of dictionary databases used to translate the entity references, improves the accuracy of $\sim60\%$ as shown in \Cref{fig:generic}. This result indicates that the more useful data is available for a specific task, the more accurate the model will be in the prediction.

\section{CONCLUSIONS}
\label{sec:discussion}

The results presented in this work allow us to conclude that, in the case of \glspl{kge}, what most influences a prediction task is the graph's structure. Our methodology also significantly reduces the computational resources necessary to train the model and produce excellent results in the specific \gls{dr} task. The embedding model helps to understand which parts of the graph are essential for a specific task and suggests which parts improve. 
Possible future works concern the extension of the \gls{kg} with other types of entities and the complete automation of the analysis procedure.

%\IEEEtriggeratref{13}
\bibliographystyle{IEEEtran} 
\bibliography{biblio}

\end{document}